\documentclass{article}
\usepackage{ijcai23}

\usepackage{times}
\usepackage{soul}
\usepackage{url}
\usepackage[hidelinks]{hyperref}
\usepackage[utf8]{inputenc}
\usepackage[small]{caption}
\usepackage{graphicx}
\usepackage{amsmath}
\usepackage{amsthm}
\usepackage{booktabs}
\usepackage{algorithm}
\usepackage{algorithmic}
\usepackage{multirow}
\usepackage{enumitem}
\usepackage{url,graphicx,color,xcolor,booktabs,colortbl,threeparttable}
\usepackage[switch]{lineno}

\title{DAMO-StreamNet: Optimizing Streaming Perception in Autonomous Driving}
\author{
Jun-Yan He$^1$\thanks{Denotes equal contributions}\and
Zhi-Qi Cheng$^2$\footnotemark[1]\thanks{Z. Cheng is the corresponding author}\and
Chenyang Li$^1$\footnotemark[1]\and
Wangmeng Xiang$^1$\footnotemark[1]\and \\
Binghui Chen$^1$\and
Bin Luo$^1$\and
Yifeng Geng$^1$\and
Xuansong Xie$^1$
\affiliations
$^1$DAMO Academy, Alibaba Group~~~~~~~
$^2$Carnegie Mellon University ~~~~~~~
\emails
\{leyuan.hjy, wangmeng.xwm, luwu.lb, cangyu.gyf\}@alibaba-inc.com,
zhiqic@cs.cmu.edu, 
lichenyang.scut@foxmail.com,
chenbinghui@bupt.cn,
xingtong.xxs@taobao.com}
\begin{document}

\maketitle

\begin{abstract}
Real-time perception, or streaming perception, is a crucial aspect of autonomous driving that has yet to be thoroughly explored in existing research. To address this gap, we present DAMO-StreamNet, an optimized framework that combines recent advances from the YOLO series with a comprehensive analysis of spatial and temporal perception mechanisms, delivering a cutting-edge solution. The key innovations of DAMO-StreamNet are:~(1)~A robust neck structure incorporating deformable convolution, enhancing the receptive field and feature alignment capabilities.~(2)~A dual-branch structure that integrates short-path semantic features and long-path temporal features, improving motion state prediction accuracy.~(3)~Logits-level distillation for efficient optimization, aligning the logits of teacher and student networks in semantic space.~(4)~A real-time forecasting mechanism that updates support frame features with the current frame, ensuring seamless streaming perception during inference.~Our experiments demonstrate that DAMO-StreamNet surpasses existing state-of-the-art methods, achieving 37.8\% (normal size (600, 960)) and 43.3\% (large size (1200, 1920)) sAP without using extra data. This work not only sets a new benchmark for real-time perception but also provides valuable insights for future research. Additionally, DAMO-StreamNet can be applied to various autonomous systems, such as drones and robots, paving the way for real-time perception\footnote{Code is at https://github.com/zhiqic/DAMO-StreamNet}.
\end{abstract}

\begin{figure} [!ht]
\small
\begin{center}
\includegraphics[width=0.95\linewidth]{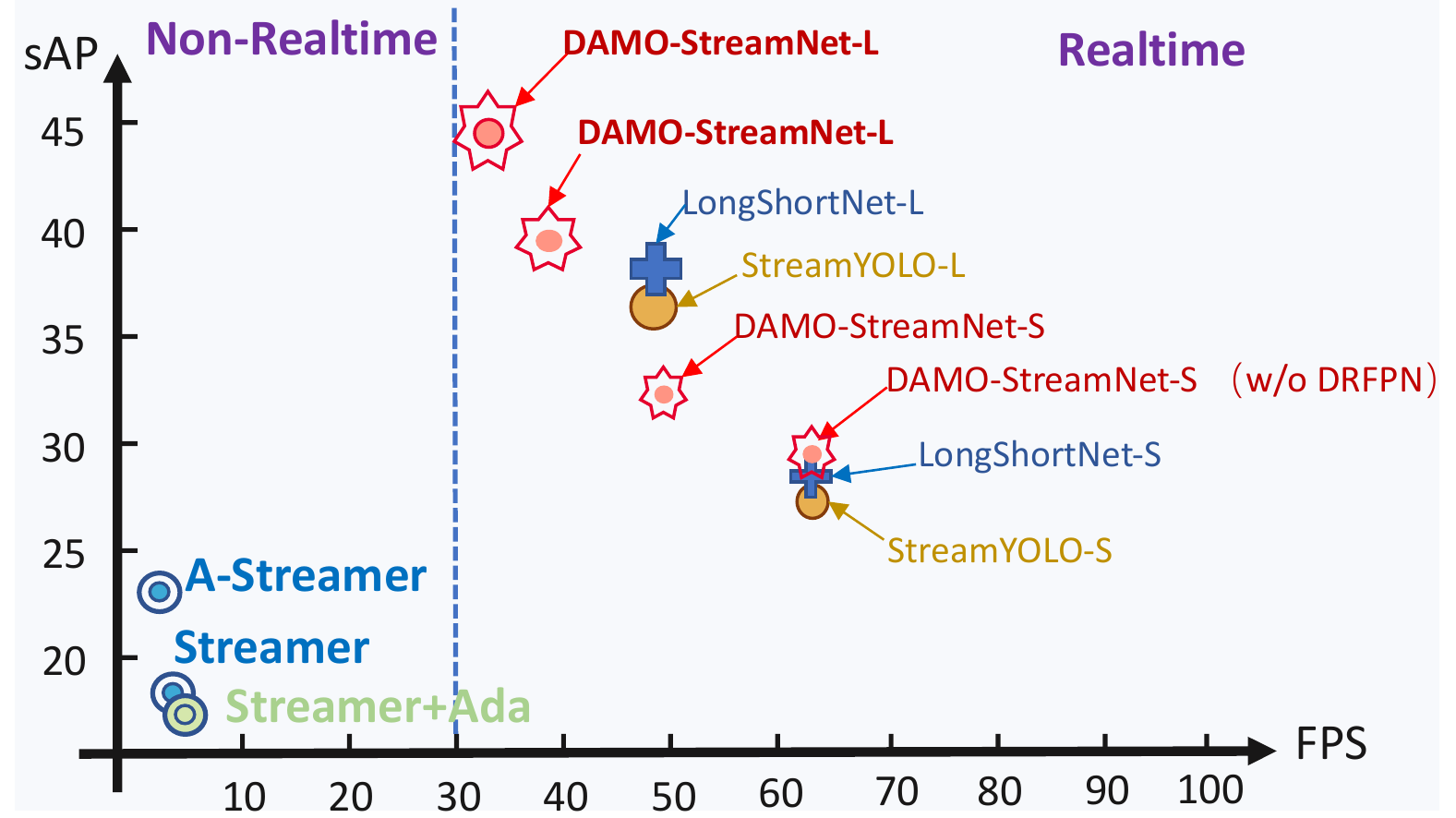}
\end{center}
\vspace{-4mm}
\caption{\small Performance comparisons of streaming perception task, showcasing the balance between accuracy and speed achieved by our proposed method, DAMO-StreamNet, which sets a new state-of-the-art benchmark.}
\label{fig:introduction}
\vspace{-4mm}
\end{figure}

\section{Introduction}
The advent of autonomous vehicles has driven the need for high-performance traffic environment perception systems. In this context, streaming perception, which involves detecting and tracking objects in a video stream simultaneously, is a fundamental technique that significantly impacts autonomous driving decision-making. Notably, the fast-changing scale of traffic objects due to vehicle motion can lead to conflicts in the receptive field when detecting both large and small objects. Moreover, real-time perception is an ill-posed problem that heavily depends on motion consistency context and historical data. Consequently, two major challenges in real-time perception are:~(1) adaptively handling rapidly changing object scales, and~(2) accurately and efficiently learning long-term motion consistency.

Despite previous research on temporal aggregation techniques~\cite{Wang_MANet_ECCV18,Chen_ScaleTime_CVPR2018,Lin_DualSem_MM20,SunHHR21_MAMBA_AAAI,Huang_TAda_ICLR22} has primarily focused on offline settings and is unsuitable for online real-time perception. Furthermore, enhancing the base detector has not been thoroughly investigated in the context of real-time perception. To address these limitations, we propose DAMO-StreamNet, a practical real-time perception pipeline that improves the model in four key aspects:

\begin{enumerate}
\item \textit{To enhance the base detector's performance}, we propose an efficient feature aggregation scheme called Dynamic Receptive Field FPN. This scheme utilizes connections and deformable convolution networks to resolve receptive field conflicts and bolster feature alignment capacity. We also adopt the cutting-edge detection technique Re-parameterized to further enhance the network's performance without adding extra inference costs. \textit{These improvements lead to higher detection accuracy and faster inference times.}

\item \textit{To capture long-term spatial-temporal correlations}, we design a dual-path structure temporal fusion module. \textit{This module employs a two-stream architecture that separates spatial and temporal information, facilitating the accurate and efficient capture of long-term correlations.}

\item \textit{To tackle the challenges of learning long-term motion consistency}, we propose an Asymmetric Knowledge Distillation (AK-Distillation) framework. This framework employs a teacher-student learning strategy in which student networks are supervised by transferring the generalized knowledge captured by large-scale teacher networks. \textit{This method enforces the long-term motion consistency of the feature representations between the teacher-student pair, resulting in enhanced performance.}

\item \textit{To fulfill the real-time forecasting requirement}, we update the support frame features with the current frame before the next prediction in the inference phase. Additionally, the support frame features are updated by the current frame to prepare for the next prediction in the inference phase to satisfy the real-time forecasting requirement. \textit{This approach allows for the pipeline to handle real-time streaming perception and make predictions on time.}
\end{enumerate}

\textit{In summary, DAMO-StreamNet offers a state-of-the-art solution for real-time perception in autonomous driving.} We also introduce a new evaluation metric, the K-Step Streaming Metric, which considers the temporal interval to assess real-time perception. Our experiments show that DAMO-StreamNet outperforms existing SOTA methods, achieving 37.8\% (normal size (600, 960)) and 43.3\% (large size (1200, 1920)) sAP without using any extra data. Our work not only establishes a new benchmark for real-time perception but also provides valuable insights for future research in this field. Moreover, DAMO-StreamNet can be applied to various types of autonomous systems, such as drones and robots, enabling real-time and accurate environmental perception. This, in turn, can enhance the safety and efficiency of these systems, making them more practical for everyday use\footnote{Code is at https://github.com/zhiqic/DAMO-StreamNet.}.

\begin{figure} [!ht]
\small
\begin{center}
\includegraphics[width=0.8\linewidth]{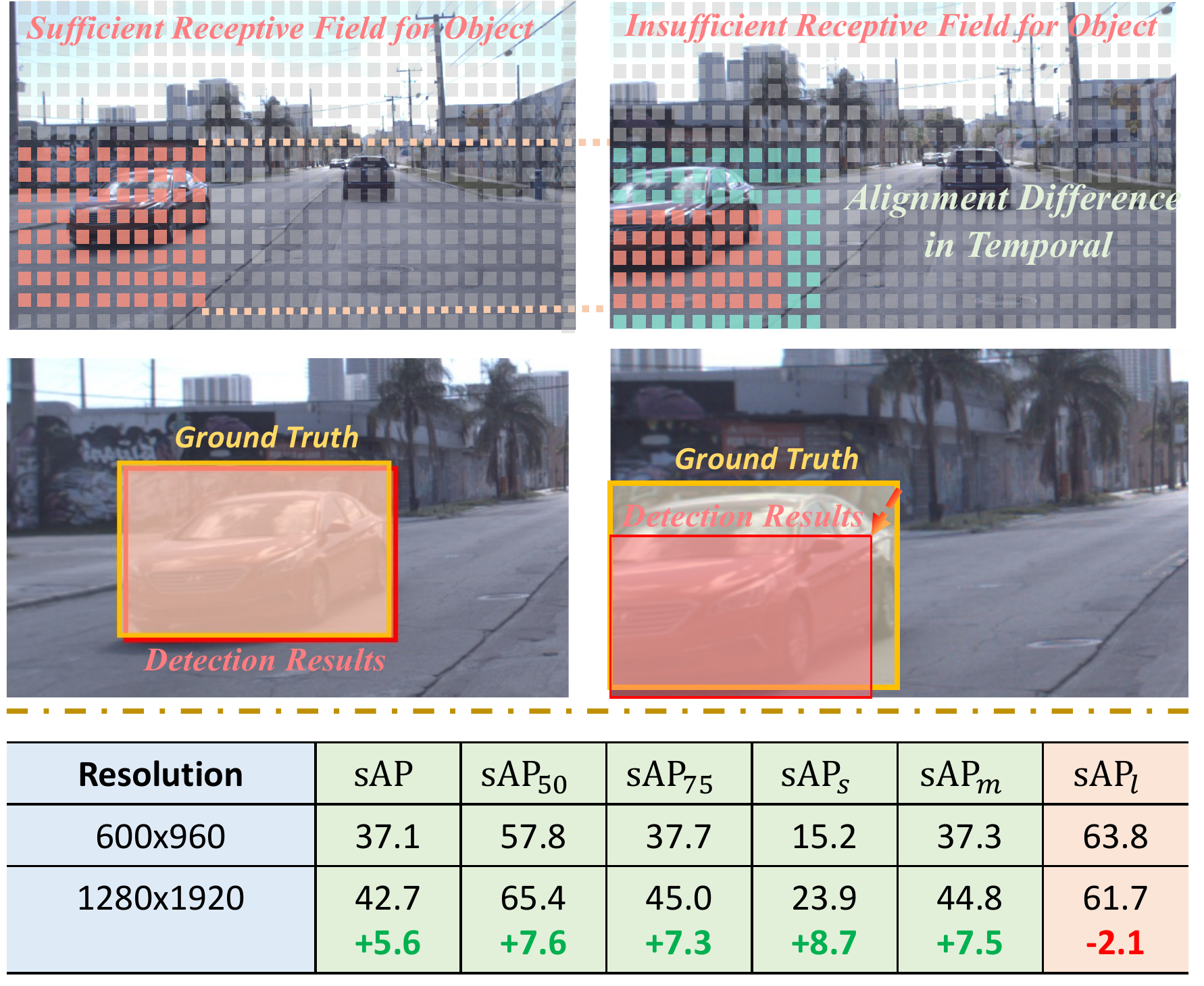}
\end{center}
\vspace{-4mm}
\caption{\small Impact of Receptive Field on Streaming Perception: Inadequate receptive field coverage, as illustrated in the upper and middle regions, leads to unsuccessful predictions. This finding underscores the decrease in performance for large-scale objects with high-resolution input, attributed to limited receptive field coverage.}
\label{fig:recepfield}
\vspace{-2mm}
\end{figure}

\begin{figure*}
\begin{center}
\includegraphics[width=0.95\linewidth]{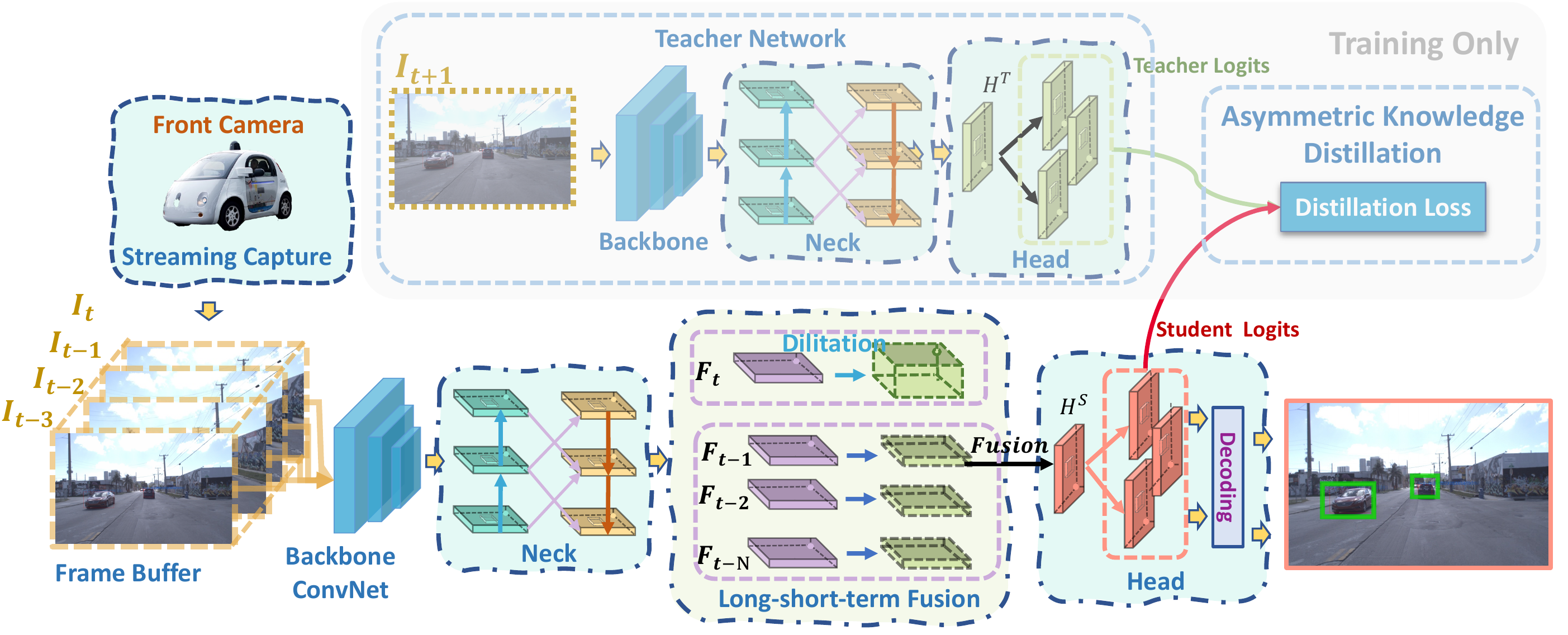}
\end{center}
\vspace{-2mm}
\caption{An overview of the proposed DAMO-StreamNet framework. The upper part, obscured by the white mask, contains the teacher network and the Asymmetric Knowledge Distillation module, which are utilized exclusively during the training phase. The lower part represents the student network, featuring the backbone, neck, long-short-term fusion module, and head for efficient streaming perception.}
\vspace{-2mm}
\label{fig:framework}
\end{figure*}

\section{Related Work}
\subsection{Image Object Detection}
\noindent \textbf{State-of-the-art~Detectors}.~In recent years, remarkable progress in deep learning-based object detection has been witnessed. Image object detection is fundamental to streaming perception. Therefore, we first review the state-of-the-art detectors~\cite{Ge_YOLOX_CORR21,Wang_YOLOv7_CORR22} and cutting-edge techniques from multiple aspects, including backbone design~\cite{Wang_HRNet_TPAMI20,Ding_DBranch_CVPR21,Ding_RepVGG_CVPR21,Ding_ACNet_ICCV19,Kumar_Mobile_CoRR22}, effective feature aggregation \cite{Lin_FPN_CVPR17,GhiasiLL_NASFPN_CVPR19,Jiang_GiraffeDet_ICLR22,Tan_EfficientDet_CVPR20,cheng2022gsrformer,tu2023implicit}, and optimal label assignment~\cite{Ge_OTA_CVPR21,Kim_PAA_ECCV20,Carion_DETR_ECCV20}.

\noindent \textbf{Feature~Aggregation}.~Associated with the backbone network development, the feature aggregation solution, FPN~\cite{Lin_FPN_CVPR17} and PAFPN \cite{Liu_PAFPN_CVPR18} are known as `necks' in the general detection pipeline. Neural Architecture Search (NAS) is also applied to this topic, introducing NAS-FPN~\cite{GhiasiLL_NASFPN_CVPR19,cheng2018learning,huang2018gnas} for object detection. All the efforts aforementioned are mainly for bridging the representation gap between classification and object detection. Beyond this setting, GiraffeDet~\cite{Jiang_GiraffeDet_ICLR22} adopts an extremely lightweight backbone but a heavy neck for feature learning.

\subsection{Video Object Detection}
\noindent \textbf{Temporal Learning}.~A common schema to learn the temporal dynamics is feature aggregation which boosts per-frame feature representation by aggregating the features of nearby frames~\cite{Wang_MANet_ECCV18,Chen_ScaleTime_CVPR2018,Lin_DualSem_MM20,SunHHR21_MAMBA_AAAI,lan2022procontext}. DeepFlow~\cite{Zhu_Deepflow_CVPR17} and FGFA~\cite{Zhu_Flowguide_ICCV17} utilize the optic flow from FlowNet \cite{Dosovitskiy_FlowNet_ICCV15} to model motion relations via different temporal feature aggregation.~MANet~\cite{Wang_MANet_ECCV18} self-adaptively combines pixel-level and instance-level calibration according to the motion in a unified framework to calibrate the features at pixel-level with inaccurate flow estimation.

\noindent \textbf{Temporal Linking}.~Despite the gratifying success of these approaches, most of the pipelines for video object detection are overly sophisticated, requiring extra temporal modeling components, e.g., optical flow model~\cite{Zhu_Deepflow_CVPR17}, recurrent neural network~\cite{Lin_DualSem_MM20,he2021db}, feature alignment module~\cite{Xiao_STAligned_ECCV18,qiao2022real,he2021mgseg}, relation networks~\cite{Gao_VidVRD_MM21}. An effective and simple way for VOD is by adopting a temporal linking module such as Seq-NMS \cite{Han_SeqNMS_CoRR16}, Tubelet rescoring~\cite{Kang_TCNN_CVPR16} and Seq-Bbox Matching~\cite{Belhassen_SeqBbox_VISAPP19,lan2022procontext} as post-processing, which links the same object across the video to form tubelets and aggregating classification scores to achieve the state-of-the-art performance.

\subsection{Knowledge Distillation}
Knowledge distillation~\cite{Hinton_distilling_2015} is designed to transfer robust feature representation from the teacher network to the student network. The most intuitive solution is to encourage the student model to learn the response of the teacher model~\cite{Hinton_distilling_2015}. TAKD~\cite{Mirzadeh_TA_AAAI2020} proposes to leverage an intermediate-sized teacher-assistant network to fill the gap between the teacher and student network. Also, Fitnets~\cite{Chen_learning_NeurIPS2017} proposes to guide the student network with the intermediate features of the teacher network. Meanwhile,~\cite{Chen_learning_NeurIPS2017} first leverages hint learning of intermediate layers in the multi-class object detection task. More works~\cite{Heo_Knowledge_AAAI2019,Chen_SemCKD_AAAI2021,cheng2018learning,cheng2019improving} design different intermediate representations to transfer knowledge of the teacher network. Moreover,~\cite{Yao_Gdetkd_CVPR2021,Liu_structured_TPAMI2020,Yang_FGD_CVPR2022,cheng2019learning} propose to learn the correct relation between different data samples or layers. As far as we know, DAMO-StreamNet is the first work to leverage knowledge distillation in the streaming perception task. The knowledge distillation module encourages the network to obtain more accurate results in ``predicting the next frame"~\cite{Yang_Streamyolo_CVPR22} by mimicking the features of the ``next frame".

\subsection{Streaming Perception}
As the streaming perception task is recently proposed, only a few works focus on it. Most methods~\cite{LiWR_Streaming_ECCV20} are object detection based and integrate with the temporal modeling techniques to boost the performance. However, the current SOTA work StreamYOLO~\cite{Yang_Streamyolo_CVPR22} wholly ignores the semantics and motion in video streams and only uses the last two frames as input. The current SOTA streaming perception work~\cite{Yang_Streamyolo_CVPR22,LongShortNet_CoRR22} is constructed based on the YOLOX-based detector. In this work, we revamp the base detector of DAMO-StreamNet for the issues derived from the state-of-the-art methods associated with re-parameter, feature aggregation, and knowledge distillation. Our approach provides a more comprehensive and sophisticated solution for the streaming perception task, addressing the limitations of existing methods and setting a new benchmark for future research.
\begin{figure*}
\begin{center}
\includegraphics[width=0.85\linewidth]{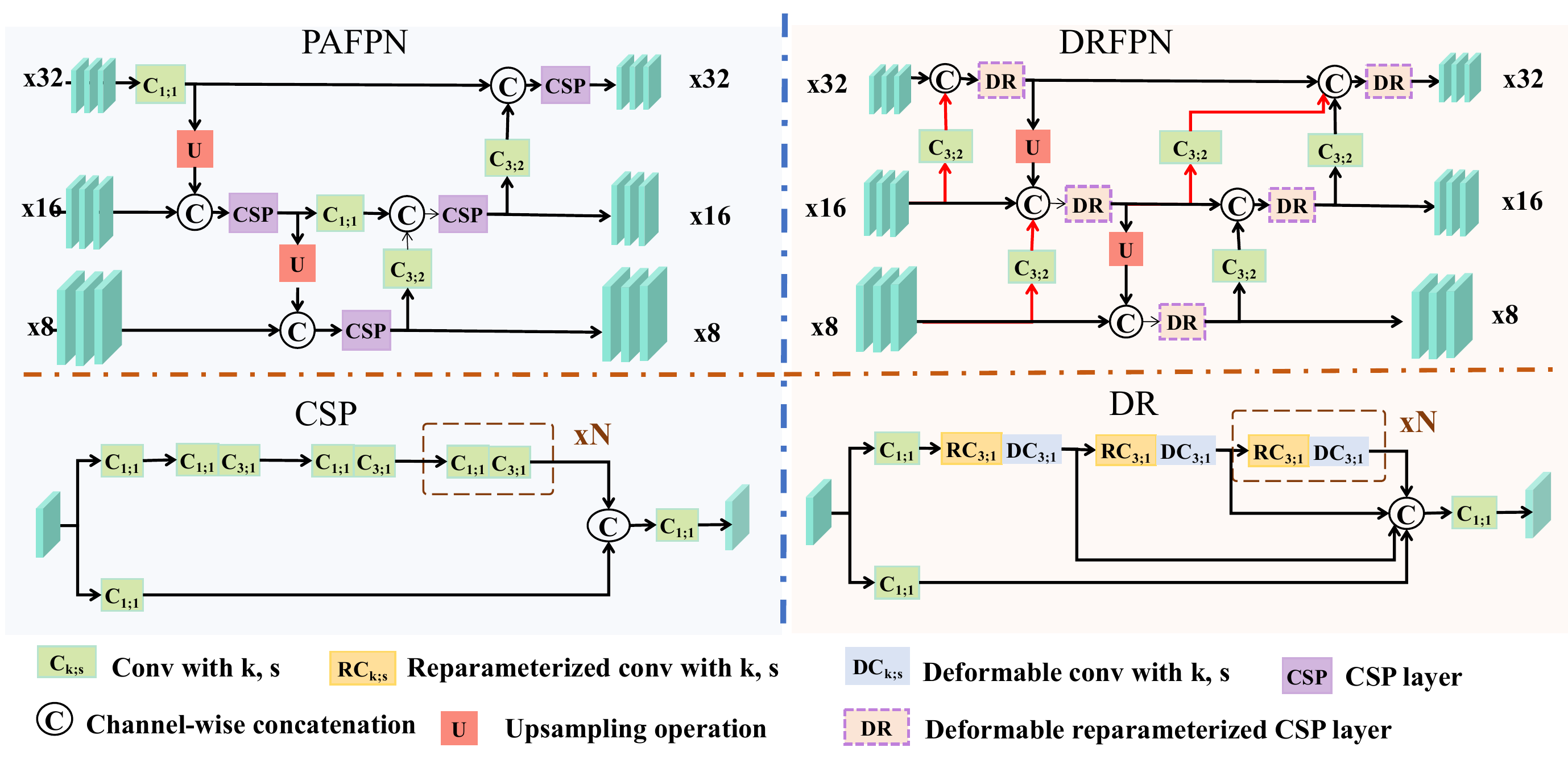}
\end{center}
\vspace{-4mm}
\caption{A comprehensive comparison between PAFPN and our proposed DRFPN, both constructed using the base block CSP and DR layer. The notation ``Conv with k, s" represents a convolution layer with kernel size `k' and stride `s'.}
\vspace{-4mm}
\label{fig:drfpn}
\end{figure*}
\section{DAMO-StreamNet}
The overall framework is illustrated in Fig.~\ref{fig:framework}. Initially, a video frame sequence passes through DAMO-StreamNet to extract spatiotemporal features and generate the final output feature. Subsequently, the Asymmetric Knowledge Distillation module (AK-Distillation) takes the output logit features of the teacher and student networks as inputs, transferring the semantics and spatial position of the future frame extracted by the teacher to the student network.

Given a video frame sequence $\mathcal{S}=\{I_{t},\dots I_{t-N\delta t}\}$, where $N$ and $\delta t$ represent the number and step size of the frame sequence, respectively. DAMO-StreamNet can be defined as,
\begin{equation*}
\centering
\mathcal{T}=\mathcal{F}(\mathcal{S}, W),
\end{equation*}
where $W$ denotes the network weights, and $\mathcal{T}$ represents the collection of final output feature maps. $\mathcal{T}$ can be further decoded using $Decode(\mathcal{T})$ to obtain the result $\mathcal{R}$, which includes the score, category, and location of the objects.

In the training phase, the student network can be represented as,
\begin{equation*}
\centering
\mathcal{T}_{stu}=\mathcal{F}_{stu}(\mathcal{S}, W_{stu}).
\end{equation*}
Besides the student network, the teacher network takes the $t+1$ frame as input to generate the future result, represented by,
\begin{equation*}
\centering
\mathcal{T}_{tea}=\mathcal{F}_{tea}(I_{t+1}, W_{tea}),
\end{equation*}
where $W_{stu}$ and $W_{tea}$ denote the weights of the student and teacher networks, respectively. Then, AK-Distillation leverages $\mathcal{T}_{stu}$ and $\mathcal{T}_{tea}$ as inputs to perform knowledge distillation $\text{AKDM}(\mathcal{T}_{stu}, \mathcal{T}_{tea})$. More details are elaborated in the following subsections.

\subsection{Network Architecture}
The network is composed of three elements: the backbone, neck, and head. It can be formulated as,
\begin{equation*}
\centering
\mathcal{T}=\mathcal{F}(\mathcal{S}, W)=\mathcal{G}_{h}(\mathcal{G}_{n}(\mathcal{G}_{b}(\mathcal{S}, W_{b}), W_{n}), W_{h}),
\end{equation*}
where $\mathcal{G}_{b}$, $\mathcal{G}_{n}$, and $\mathcal{G}_{h}$ stand for the backbone, neck, and head components respectively, while $W_{b}$, $W_{n}$, and $W_{h}$ symbolize their corresponding weights. Previous studies~\cite{Jiang_GiraffeDet_ICLR22} highlighted the neck structure's critical role in feature fusion and representation learning for detection tasks. Consequently, we introduce the Dynamic Receptive Field FPN (DRFPN), which employs a learnable receptive field approach for enhanced feature fusion. To benchmark against the current state-of-the-art (SOTA), we apply the same settings for $\mathcal{G}_{n}$, $\mathcal{G}_{h}$, and StreamYOLO~\cite{Yang_Streamyolo_CVPR22}, leveraging CSPDarknet-53~\cite{Ge_YOLOX_CORR21} and TALHead~\cite{Yang_Streamyolo_CVPR22} to build the network. Given the proven efficacy of long-term temporal information by the existing LongShortNet~\cite{LongShortNet_CoRR22}, we also integrate a dual-path architectural module for spatial-temporal feature extraction.

\noindent \textbf{Dynamic~Receptive~Field~FPN}. Recent object detection studies, including StreamYOLO~\cite{Yang_Streamyolo_CVPR22} and LongShortNet \cite{LongShortNet_CoRR22}, have utilized YOLOX as their fundamental detector. YOLOX's limitation is its fixed spatial receptive field that cannot synchronize features temporally, thus impacting its performance. To address this, we propose the Dynamic Receptive Field FPN (DRFPN) with a learnable receptive field strategy and an optimized fusion mechanism.

Specifically, Fig.\ref{fig:drfpn} contrasts PAFPN and DRFPN. PAFPN employs sequential top-down and bottom-up fusion operations to amplify feature representation. However, conventional convolution with a static kernel size fails to align features effectively. As a solution, we amalgamate the DRM module and Bottom-up Auxiliary Connect (BuAC) with PAFPN to create DRFPN. We introduce three notable modifications compared to PAFPN's CSP module (Fig.\ref{fig:drfpn}):(1) We integrate deformable convolution layers into the DRFPN module to provide the network with learnable receptive fields;(2) To enhance feature representation, we adopt re-parameterized convolutional layers \cite{Ding_RepVGG_CVPR21};(3) ELAN \cite{Wang_YOLOv7_CORR22} and Bottom-up Auxiliary Connect bridge the semantic gap between low and high-level features, ensuring effective detection of objects at diverse scales.

\noindent \textbf{Dual-Path Architecture}.~ The existing StreamYOLO \cite{Yang_Streamyolo_CVPR22} relies on a single historical frame in conjunction with the current frame to learn short-term motion consistency. While this suffices for ideal uniform linear motion, it falls short in handling complex motion, such as non-uniform motion (e.g., accelerating vehicles), non-linear motion (e.g., rotation of objects or camera), and scene occlusions (e.g., billboard or oncoming car occlusion).

To remedy this, we integrate the dual-path architecture \cite{LongShortNet_CoRR22} with a reimagined base detector, enabling the capture of long-term temporal motion while calibrating it with short-term spatial semantics. The original backbone and neck can be represented formally as,
\begin{equation*}
\centering
\begin{split}
&\mathcal{G}_{n}(\mathcal{G}_{b}(\mathcal{S}, W_{b}), W_{n})
\\&=\mathcal{G}_{n+b}(\mathcal{S}, W_{n+b})
\\&=\mathcal{G}_{fuse}(\mathcal{G}_{n+b}^{short}(I_{t}), \mathcal{G}_{n+b}^{long}(I_{t-\delta t}, \dots, I_{t-N\delta t})),
\end{split}
\end{equation*}
where $\mathcal{G}_{fuse}$ represents the LSFM-Lf-Dil of LongShortNet. $\mathcal{G}_{n+b}^{short}$ and $\mathcal{G}_{n+b}^{long}$ denote the ShortPath and LongPath of LongShortNet, which are used for feature extraction of the current and historical feature, respectively.~Note that their weights are shared.

Finally, the dual-path network is formulated as,
\begin{equation*}
\centering
\begin{split}
\mathcal{T}&=\mathcal{F}(\mathcal{S}, W)
\\&=\mathcal{G}_{h}(\mathcal{G}_{n}(\mathcal{G}_{b}(\mathcal{S}, W_{b}), W_{n}), W_{h})
\\&=\mathcal{G}_{h}(\mathcal{G}_{fuse}(\mathcal{G}_{n+b}^{short}(I_{t}), \mathcal{G}_{n+b}^{long}(I_{t-\delta t}, \dots, I_{t-N\delta t}))),
\end{split}
\end{equation*}
where the proposed dual-path architecture effectively addresses complex motion scenarios and offers a sophisticated solution for object detection in video sequences.

\subsection{Asymmetric Knowledge Distillation}
The ability to retain long-term spatiotemporal knowledge through fused features lends strength to forecasting, yet achieving streaming perception remains a daunting task. Drawing inspiration from knowledge distillation, we've fashioned an asymmetric distillation strategy, transferring ``future knowledge" to the present frame. This assists the model in honing its accuracy in streaming perception without the burden of additional inference costs.

Given the asymmetric input nature of the teacher and student networks, a sizable gap emerges in their feature distributions, thus impairing the effectiveness of distillation at the feature level. Logits-based distillation primarily garners performance improvements by harmonizing the teacher model's response-based knowledge, which aligns knowledge distribution at the semantic level. This simplifies the optimization process for asymmetric distillation. As a result, we've engineered a distillation module to convey rich semantic and localization knowledge from the teacher (the future) to the student (the present).

The asymmetric distillation is depicted in Fig.~\ref{fig:framework}. The teacher model is a still image detector that takes $I_{t+1}$ as input and produces logits for $I_{t+1}$. The student model is a standard streaming perception pipeline that uses historical frames {$I_{t-1}$, $\dots$, $I_{t-N}$} and the current frame $I_{t}$ as input to forecast the results of the arriving frame $I_{t+1}$. The logits produced by the teacher and student are represented by $\mathcal{T}_{stu}=\{F_{stu}^{cls}, F_{stu}^{reg}, F_{stu}^{obj}\}$, and $\mathcal{T}_{tea}=\{F_{tea}^{cls}, F_{tea}^{reg}, F_{tea}^{obj}\}$, where $F_{\cdot}^{cls}$, $F_{\cdot}^{reg}$, and $F_{\cdot}^{obj}$ correspond to the classification, objectness, and regression logits features, respectively. The Asymmetric Knowledge Distillation, $\text{AKDM}(\cdot)$, is mathematically formulated as,
\begin{equation*}
\centering
\begin{split}
&\text{AKDM}(\mathcal{T}_{stu}, \mathcal{T}_{tea})
\\&=\mathcal{L}_{cls}(F_{stu}^{cls}, F_{tea}^{cls}) + \mathcal{L}_{obj} (F_{stu}^{obj}, F_{tea}^{obj}) + \mathcal{L}_{reg}(\hat{F}_{stu}^{reg}, \hat{F}_{tea}^{reg}),
\end{split}
\vspace{-0.1in}
\end{equation*}
where $\mathcal{L}_{cls}(\cdot)$ and $\mathcal{L}_{obj}(\cdot)$ are Mean Square Error (MSE) loss functions, and $\mathcal{L}_{reg}(\cdot)$ is the GIoU loss \cite{Rezatofighi_GIOU_CVPR19}. $\hat{F}_{stu}^{reg}$ and $\hat{F}_{tea}^{reg}$ represent the positive samples of the regression logit features, filtered using the OTA assignment method as in YOLOX \cite{Ge_YOLOX_CORR21}. It is worth noting that location knowledge distillation is only performed on positive samples to avoid noise from negative ones.

\subsection{K-step Streaming Metric}
The Streaming Average Precision (sAP) metric is a prevalent tool used to gauge the precision of Streaming Perception systems \cite{LiWR_Streaming_ECCV20}. This metric gauges precision by juxtaposing real-world ground truth with system-generated results, factoring in process latency.

\begin{figure}[t]
\centering
\centerline{\includegraphics[width=0.95\linewidth]{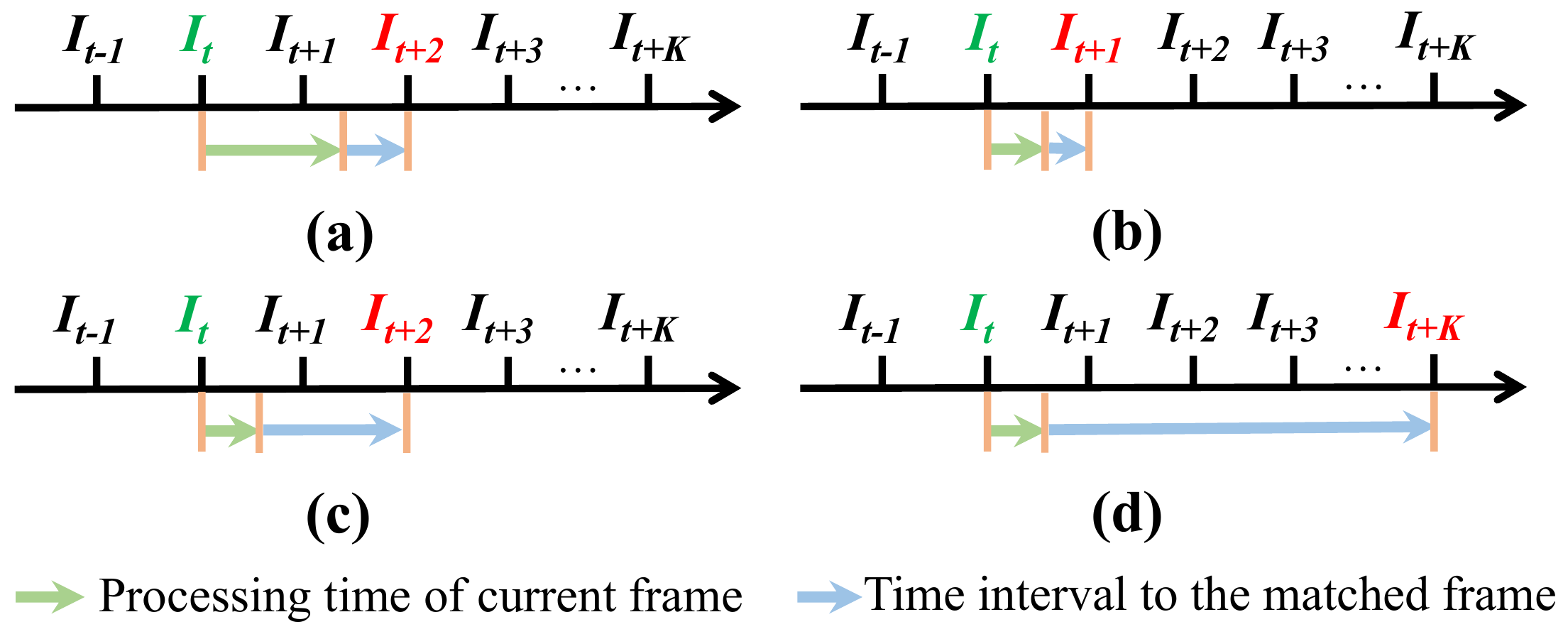}}
\vspace{-2mm}
\caption{Illustration of matching rules under different metrics. The frames in \textcolor[RGB]{0,153,51}{green} font denote the current frame and the frames in \textcolor[RGB]{255,100,97}{red} font denote the frames matched with the current frame under the specific metric. (a) Matching result of non-real-time methods under 1-sAP. (b) Matching result of real-time methods under 1-sAP. (c) Matching result of real-time methods under 2-sAP. (d) Matching result of real-time methods under K-sAP.}
\label{fig:metric}
\vspace{-4mm}
\end{figure}

Two primary methodologies exist in this domain: non-real-time and real-time. For non-real-time methods, as depicted in Fig.\ref{fig:metric}(a), the sAP metric calculates precision by comparing the current frame $I_{t}$ results with the ground truth of the following frame $I_{t+2}$, post processing of frame $I_{t}$. Conversely, real-time methods, as demonstrated in Fig.~\ref{fig:metric}(b), conclude the processing of the current frame $I_{t}$ prior to the next frame $I_{t+1}$ arrival. Our proposed method, DAMO-StreamNet, is a real-time method, adhering to the pipeline outlined in Fig.~\ref{fig:metric}(b).

Though the sAP metric effectively evaluates the short-term forecasting capability of algorithms, it falls short in assessing their long-term forecasting prowess— a critical factor in real-world autonomous driving scenarios. In response, we introduce the K-step Streaming metric, an expansion of the sAP metric, specifically tailored to evaluate long-term performance. As depicted in Fig.~\ref{fig:metric}(c), the algorithm projects the results of the upcoming two frames, and the cycle continues. The projection of the next K frames is represented as "K-sAP", as shown in Fig.~\ref{fig:metric}(d). Consequently, the standard sAP metric translates to 1-sAP in the K-step metric context.

\begin{table*}[ht] 
\footnotesize
	\centering
	\label{tab:comparing-sota}
 	\caption{Comparison with both non-real-time and real-time state-of-the-art (SOTA) methods on the Argoverse-HD benchmark dataset. The symbol '$^{\ddagger}$' denotes the use of a large size (1200, 1920) and extra data. The symbol '$^{\dagger}$' denotes the use of a large size (1200, 1920) without the use of extra data. The best results for each setting are shown in \textcolor[RGB]{0,153,51}{green}. The largest increments of the large resolution setting are shown in \textcolor[RGB]{255,100,97}{red}.}
     \vspace{-2mm}
	\setlength{\tabcolsep}{1.25mm}{
		\begin{tabular}{c|c c c|c c c}
			\hline \hline
			Methods & sAP & sAP$_{50}$ & sAP$_{75}$ & sAP$_{s}$ & sAP$_{m}$ & sAP$_{l}$ \\ \hline
			\multicolumn{7}{ c }{Non-real-time detector-based methods} \\ \hline
			Streamer (S=900) \cite{LiWR_Streaming_ECCV20} & 18.2 & 35.3 & 16.8 & \textbf{\textcolor[RGB]{0,153,51}{4.7}} & 14.4 & 34.6 \\
			Streamer (S=600) \cite{LiWR_Streaming_ECCV20} & 20.4 & 35.6 & 20.8 & 3.6 & 18.0 & 47.2 \\ 
			Streamer + AdaScale \cite{Chin_AdaScale_MLSys,Ghosh_Adaptive_CoRR21} & 13.8 & 23.4 & 14.2 & 0.2 & 9.0 & 39.9 \\ 
			Adaptive Streamer \cite{Ghosh_Adaptive_CoRR21}   & \textbf{\textcolor[RGB]{0,153,51}{21.3}} & \textbf{\textcolor[RGB]{0,153,51}{37.3}} & \textbf{\textcolor[RGB]{0,153,51}{21.1}} & 4.4 & \textbf{\textcolor[RGB]{0,153,51}{18.7}} & \textbf{\textcolor[RGB]{0,153,51}{47.1}} \\ \hline
			\multicolumn{7}{ c }{Real-time detector-based methods} \\ \hline
			StreamYOLO-S \cite{Yang_Streamyolo_CVPR22} & 28.8 & 50.3 & 27.6 & 9.7 & 30.7 & 53.1 \\
			StreamYOLO-M \cite{Yang_Streamyolo_CVPR22} & 32.9 & 54.0 & 32.5 & 12.4 & 34.8 & 58.1 \\ 
			StreamYOLO-L \cite{Yang_Streamyolo_CVPR22} & 36.1 & 57.6 & 35.6 & 13.8 & 37.1 & 63.3 \\
            \hline 
			LongShortNet-S \cite{LongShortNet_CoRR22} & 29.8 & 50.4 & 29.5 & 11.0 & 30.6 & 52.8 \\
			LongShortNet-M \cite{LongShortNet_CoRR22} & 34.1 & 54.8 & 34.6 & 13.3 & 35.3 & 58.1 \\ 
			LongShortNet-L \cite{LongShortNet_CoRR22} & 37.1 & 57.8 & 37.7 & 15.2 & 37.3 & 63.8 \\
            \hline 
			DAMO-StreamNetNet-S (Ours)     & 31.8 & 52.3 & 31.0 & 11.4 & 32.9 & 58.7 \\ 
			DAMO-StreamNetNet-M (Ours)    & 35.7 & 56.7 & 35.9 & 14.5 & 36.3 & 63.3 \\ 
			DAMO-StreamNetNet-L (Ours)     & \textbf{\textcolor[RGB]{0,153,51}{37.8}}  & \textbf{\textcolor[RGB]{0,153,51}{59.1}} & \textbf{\textcolor[RGB]{0,153,51}{38.6}} & \textbf{\textcolor[RGB]{0,153,51}{16.1}} & \textbf{\textcolor[RGB]{0,153,51}{39.0}} & \textbf{\textcolor[RGB]{0,153,51}{64.6}} \\ 
			\hline
			\multicolumn{7}{ c }{Large resolution} \\ \hline
			StreamYOLO-L $^{\ddagger}$    & 41.6 & 65.2 & 43.8 & 23.1 & 44.7 & 60.5 \\ 
			  LongShortNet-L $^{\dagger}$     & 42.7 (+1.1) & 65.4 (+0.2) & \textbf{\textcolor[RGB]{0,153,51}{45.0}} (\textbf{\textcolor[RGB]{255,100,97}{+1.2}}) & 23.9 (+0.8) & 44.8 (+0.1) & 61.7 (+1.2) \\ 
			DAMO-StreamNet-L $^{\dagger}$ (Ours)     & \textbf{\textcolor[RGB]{0,153,51}{43.3}} (\textbf{\textcolor[RGB]{255,100,97}{+1.7}})  & \textbf{\textcolor[RGB]{0,153,51}{66.1}} (\textbf{\textcolor[RGB]{255,100,97}{+0.9}}) & 44.6 (+0.8) & \textbf{\textcolor[RGB]{0,153,51}{24.2}} (\textbf{\textcolor[RGB]{255,100,97}{+1.1}}) & \textbf{\textcolor[RGB]{0,153,51}{47.3}} (\textbf{\textcolor[RGB]{255,100,97}{+2.6}}) & \textbf{\textcolor[RGB]{0,153,51}{64.1}} (\textbf{\textcolor[RGB]{255,100,97}{+3.6}}) \\ 
			\hline
	\end{tabular}}
 \vspace{-2mm}
\end{table*}

\section{Experiments}
\vspace{1mm}
\subsection{Dataset and Metric}
\noindent \textbf{Dataset}:~We utilized the Argoverse-HD dataset, which comprises various urban outdoor scenes from two US cities. The dataset contains detection annotations and center RGB camera images, which were used in our experiments. We adhered to the train/validation split proposed by Li et al.~\cite{LiWR_Streaming_ECCV20}, with the validation set consisting of 15k frames.

\noindent \textbf{Evaluation Metrics}:~We employed the streaming Average Precision (sAP) metric to evaluate performance. The sAP metric calculates the average mAP over Intersection over Union (IoU) thresholds ranging from 0.5 to 0.95, as well as APs, APm, and APl for small, medium, and large objects, respectively. This metric has been widely used in object detection, including in previous works such as~\cite{LiWR_Streaming_ECCV20,Yang_Streamyolo_CVPR22}.

\subsection{Implementation Details}
We pretrained the base detector of our DAMO-StreamNet on the COCO dataset~\cite{Lin_COCO_ECCV14}, following the methodology of StreamYOLO~\cite{Yang_Streamyolo_CVPR22}. We then trained DAMO-StreamNet on the Argoverse-HD dataset for 8 epochs with a batch size of 32, using 4 V100 GPUs. For convenient comparison with recent state-of-the-art models~\cite{Yang_Streamyolo_CVPR22,LongShortNet_CoRR22}, we designed small, medium, and large networks (i.e., DAMO-StreamNet-S, DAMO-StreamNet-M, and DAMO-StreamNet-L). The normal input resolution (600, 960) was utilized unless specified otherwise. We maintained consistency with other hyperparameters from previous works~\cite{Yang_Streamyolo_CVPR22,LongShortNet_CoRR22}. AK-Distillation is an auxiliary loss for DAMO-StreamNet training, with the weight of the loss set to 0.2/0.2/0.1 for DAMO-StreamNet-S/M/L, respectively.

\subsection{Comparison with State-of-the-art Methods}
We compared our proposed approach with state-of-the-art methods to evaluate its performance. In this subsection, we directly copied the reported performance from their original papers as their results. The performance comparison was conducted on the Argoverse-HD dataset~\cite{LiWR_Streaming_ECCV20}. An overview of the results reveals that our proposed DAMO-StreamNet with an input resolution of 600 $\times$ 960 achieves 37.8\% sAP, outperforming the current state-of-the-art methods by a significant margin. For the large-resolution input of 1200 $\times$ 1920, our DAMO-StreamNet attains 43.3\% sAP without extra training data, surpassing the state-of-the-art work StreamYOLO, which was trained with large-scale auxiliary datasets. This clearly demonstrates the effectiveness of the systematic improvements in DAMO-StreamNet.

Compared to StreamYOLO and LongShortNet, DAMO-StreamNet-L achieves absolute improvements of 3.6\% and 2.4\% under the sAP${L}$ metric, respectively. This also provides substantial evidence that the features produced by DRFPN offer a self-adaptive and sufficient size of the receptive field for large-sized objects. It is worth noting that DAMO-StreamNet experiences a slight decline compared to LongShortNet under the stricter metric sAP${75}$. This observation suggests that although the dynamic receptive field achieves a sufficient receptive field for different scales of objects, it is not as accurate as fixed kernel-size ConvNets. The offset prediction in the deformable convolution layer may not be precise enough for high-precision scenarios. In other words, better performance could be achieved if this issue is addressed, and we leave this for future work.

\begin{table}[h]
\small
    \begin{center} 
    \caption{Ablation study of the base detector on the Argoverse-HD dataset. The best results for each subset and the corresponding increments are shown in \textcolor[RGB]{0,153,51}{green} font and \textcolor[RGB]{255,100,97}{red} font, respectively.}
    \vspace{-2mm}
	\setlength{\tabcolsep}{1.35mm}{
		\begin{tabular}{c|c c c}
			\hline \hline
			Methods & S & M & L \\ \hline
            \multicolumn{4}{ c }{Equip StreamYOLO with our DRFPN} \\ \hline
			StreamYOLO                           & 28.7          & 33.5          & 36.1             \\ \hline
			+DRFPN                          & \textbf{\textcolor[RGB]{0,153,51}{30.6}} (\textbf{\textcolor[RGB]{255,100,97}{+1.9}})          & \textbf{\textcolor[RGB]{0,153,51}{35.1}} (\textbf{\textcolor[RGB]{255,100,97}{+1.6}})          & \textbf{\textcolor[RGB]{0,153,51}{36.7}} (\textbf{\textcolor[RGB]{255,100,97}{+0.6}})                          \\ \hline
            \multicolumn{4}{ c }{LongShortNet Equipped with our DRFPN} \\ \hline
			LongShortNet                           & 29.8          & 34.0          &  36.7             \\ \hline
			+DRFPN                          & \textbf{\textcolor[RGB]{0,153,51}{31.5}} (\textbf{\textcolor[RGB]{255,100,97}{+1.7}})         & \textbf{\textcolor[RGB]{0,153,51}{35.7}} (\textbf{\textcolor[RGB]{255,100,97}{+1.7}})          & \textbf{\textcolor[RGB]{0,153,51}{37.5}} (\textbf{\textcolor[RGB]{255,100,97}{+0.8}})                          \\ \hline
	\end{tabular}}
	\label{table:base-detector}
    \end{center} 
\vspace{-2mm}
\end{table}

\subsection{Ablation Study}
\vspace{1mm}
\noindent \textbf{Investigation of DRFPN.} To verify the effectiveness of DRFPN, we use StreamYOLO~\cite{Yang_Streamyolo_CVPR22} and LongShortNet~\cite{LongShortNet_CoRR22} as baselines and integrate them with the proposed DRFPN, respectively. The experimental results are listed in Table~\ref{table:base-detector}. It is evident that DRFPN significantly improves the feature aggregation capability of the baselines. Particularly, the small-scale baseline models equipped with DRFPN achieve improvements of 1.9\% and 1.7\%, separately. This also demonstrates that the dynamic receptive field is crucial for the stream perception task. More importantly, DRFPN enhances the performance of LongShortNet, which suggests that the temporal feature alignment capacity is also augmented by the dynamic receptive field mechanism.

\begin{table}[ht]
\small
\centering
   \caption{Exploration of $N$ and $\delta t$ on the Argoverse-HD dataset. StreamNet denotes our DAMO-StreamNet. The best two results and the worst one are shown in \textcolor[RGB]{0,153,51}{green} font, \textcolor[RGB]{0,0,255}{blue} font, and \textcolor[RGB]{138,43,226}{purple} font, respectively. The best increments are shown in \textcolor[RGB]{255,100,97}{red} font.}
   \vspace{-2mm}
	\setlength{\tabcolsep}{1.35mm}{
		\begin{tabular}{c|c c c}
			\hline \hline
			\textbf{($N$, $\delta t$)} & StreamNet-S & StreamNet-M & StreamNet-L \\ \hline
			(0, -)                           & \textcolor[RGB]{138,43,226}{28.1}          & \textcolor[RGB]{138,43,226}{32.0}          & \textcolor[RGB]{138,43,226}{34.2}                          \\ \hline
			(1, 1)                           & 30.6          & 35.1          & 36.7                          \\ \hline
			(1, 2)                           & 31.2          & 34.5          & 37.1             \\ \hline
			(2, 1)                           & 31.2          & \textbf{\textcolor[RGB]{0,153,51}{35.7}} (\textbf{\textcolor[RGB]{255,100,97}{+3.7}})          & \textbf{\textcolor[RGB]{0,153,51}{37.5}} (\textbf{\textcolor[RGB]{255,100,97}{+3.3}})             \\ \hline
			(2, 2)                           & \textcolor[RGB]{0,0,255}{31.4} (+3.3)          & \textcolor[RGB]{0,0,255}{35.4} (+3.4)          & 37.2             \\ \hline
			(3, 1)                           & \textbf{\textcolor[RGB]{0,153,51}{31.5}} (\textbf{\textcolor[RGB]{255,100,97}{+3.4}})          & 35.3          & 37.2             \\ \hline
			(3, 2)                           & 31.2          & 35.1          & \textcolor[RGB]{0,0,255}{37.4} (+3.2)             \\ \hline
			(4, 1)                           & 31.1          & 35.0          & 37.1            \\ \hline
			(4, 2)                           & 30.7          & 35.2          & 36.5                          \\ \hline
			(5, 1)                           & 31.1          & 35.0          & \textbf{\textcolor[RGB]{0,153,51}{37.5}} (\textbf{\textcolor[RGB]{255,100,97}{+3.3}})             \\ \hline
			(5, 2)                           & 30.9          & 34.7          & 36.9             \\ \hline
	\end{tabular}}
	\label{table:temporal-range}
\end{table}

\noindent \textbf{Investigation of Temporal Range.}
To isolate the influence of temporal range, we conduct an ablation study on $N$ and $\delta t$, as listed in Table~\ref{table:temporal-range}. (0, -) represents the model utilizing only the current frame as input. It is evident that increasing the number of input frames can enhance the model's performance, with the best results obtained when $N$ is equal to 2, 2, and 3 for DAMO-StreamNet-S/M/L, respectively. However, as the number of input frames continues to increase, the performance experiences significant declines. Intuitively, longer temporal information should be more conducive to forecasting, but the effective utilization of long-term temporal information remains a critical challenge worth investigating.

\begin{table}[ht]
\small
        \centering
        \caption{Ablation study of our proposed models. D-SN and AK-D represent DAMO-StreamNet and AK-Distillation, respectively. The best results and the largest increments are shown in \textcolor[RGB]{0,153,51}{green} font and \textcolor[RGB]{255,100,97}{red} font, respectively.}
        \vspace{-2mm}
	\setlength{\tabcolsep}{1.35mm}{
		\begin{tabular}{c|c c c}
			\hline \hline
			Methods & S & M & L \\ \hline
			D-SN (N=1)                           & 30.6          & 35.1          & 36.7             \\ \hline
   			D-SN (N=1)+AK-D                           & 31.5 (\textbf{\textcolor[RGB]{255,100,97}{+0.9}})          & 35.3 (\textbf{\textcolor[RGB]{255,100,97}{+0.2}})          & 37.1 (\textbf{\textcolor[RGB]{255,100,97}{+0.4}})                          \\ \hline
			D-SN (N=2/3)                        & 31.5          & \textbf{\textcolor[RGB]{0,153,51}{35.7}}          & 37.5                          \\ \hline
			D-SN (N=2/3)+AK-D                          & \textbf{\textcolor[RGB]{0,153,51}{31.8}} (+0.3)          & 35.5 (-0.2)          & \textbf{\textcolor[RGB]{0,153,51}{37.8}} (+0.3)                           \\ \hline
	\end{tabular}}
	\label{table:distillation}
\end{table}

\noindent \textbf{Investigation of AK-Distillation.}~AK-Distillation is a cost-free approach for enhancing the streaming perception pipeline, and we examine its impact. We perform AK-Distillation with various lengths of temporal modeling and scales of DAMO-StreamNet. As the results listed in Table~\ref{table:distillation} indicate, AK-Distillation yields improvements of 0.2\% to 0.9\% for the DAMO-StreamNet configured with $N=1$ short-term temporal modeling. This demonstrates that AK-Distillation can effectively transfer "future knowledge" from the teacher to the student. For the DAMO-StreamNet with the setting of $N=3$, AK-Distillation improves DAMO-StreamNet-S/L by only 0.3\%, but results in a slight decline for the medium-scale model. The limited improvement for long-term DAMO-StreamNet is due to the narrow performance gap between the teacher and student, and the relatively high precision is difficult to further enhance.

\begin{table}[!ht]
\small
        \centering
        	\caption{Exploration study of K-sAP on the Argoverse-HD dataset. Here, our proposed model DAMO-StreamNet is denoted as StreamNet. The best results and largest increments for each subset are shown in \textcolor[RGB]{0,153,51}{green} and \textcolor[RGB]{255,100,97}{red} font, respectively.}
         \vspace{-2mm}
	\setlength{\tabcolsep}{1.35mm}{
		\begin{tabular}{c|c|c|c}
			\hline \hline
			\multicolumn{2}{c|}{K-Step Metric} & StreamNet (N=1) & StreamNet (N=2/3)  \\ \hline
			\multirow{6}{*}{S}               & sAP$_{1}$          & 30.6       & \textbf{\textcolor[RGB]{0,153,51}{31.5}} (+0.9)      \\ \cline{2-4}
			                                   & sAP$_{2}$          & 28.3       & 29.8 (\textbf{\textcolor[RGB]{255,100,97}{+1.5}})     \\ \cline{2-4}
			                                   & sAP$_{3}$          & 24.9       & 25.9 (+1.0)     \\ \cline{2-4}
                                                  & sAP$_{4}$          & 22.1       & 23.3 (+1.2)     \\ \cline{2-4}
                                                  & sAP$_{5}$          & 21.0       & 21.8 (+0.8)     \\ \cline{2-4}
                                                  & sAP$_{6}$          & 18.8       & 20.0 (+1.2)     \\ \hline
			\multirow{6}{*}{M}               & sAP$_{1}$          & 35.1       & \textbf{\textcolor[RGB]{0,153,51}{35.7}} (+0.6)      \\ \cline{2-4}
			                                   & sAP$_{2}$          & 31.9       & 32.8 (\textbf{\textcolor[RGB]{255,100,97}{+0.9}})     \\ \cline{2-4}
			                                   & sAP$_{3}$          & 28.8       & 29.2 (+0.4)     \\ \cline{2-4}
                                                  & sAP$_{4}$          & 25.7       & 25.9 (+0.2)     \\ \cline{2-4}
                                                  & sAP$_{5}$          & 23.2       & 23.4 (+0.2)     \\ \cline{2-4}
                                                  & sAP$_{6}$          & 21.5       & 22.0 (+0.5)     \\ \hline
			\multirow{6}{*}{L}               & sAP$_{1}$          & 36.7       & \textbf{\textcolor[RGB]{0,153,51}{37.5}} (\textbf{\textcolor[RGB]{255,100,97}{+0.8}})      \\ \cline{2-4}
			                                   & sAP$_{2}$          & 33.2       & 33.9 (+0.7)     \\ \cline{2-4}
			                                   & sAP$_{3}$          & 29.8       & 30.6 (\textbf{\textcolor[RGB]{255,100,97}{+0.8}})     \\ \cline{2-4}
                                                  & sAP$_{4}$          & 27.1       & 27.2 (+0.1)     \\ \cline{2-4}
                                                  & sAP$_{5}$          & 24.2       & 25.0 (\textbf{\textcolor[RGB]{255,100,97}{+0.8}})     \\ \cline{2-4}
                                                  & sAP$_{6}$          & 22.3       & 22.7 (+0.4)     \\ \hline
	\end{tabular}}

	\label{table:metric}
\end{table}

\noindent \textbf{Investigation of K-step Streaming Metric}.~We evaluate DAMO-StreamNet with settings $N=1$ and $N=2/3$ under the new metric sAP$_{k}$, where $k$ ranges from 1 to 6. The results are listed in Table~\ref{table:metric}. It is clear that the performance progressively declines as $k$ increases, which also highlights the challenge of long-term forecasting. Another observation is that the longer time-series information leads to better performance under the new metric.

\noindent \textbf{Inference Efficiency Analysis}.~Although the proposed DRFPN has a more complex structure compared to PAFPN, DAMO-StreamNet still maintains real-time streaming perception capabilities.~For long-term fusion, we adopt the buffer mechanism from StreamYOLO~\cite{Yang_Streamyolo_CVPR22}, which incurs only minimal additional computational cost for multi-frame feature fusion.

\begin{table}[h]
\small
        \centering
        \caption{Ablation study of inference time (ms) on V100.}
        \vspace{-3mm}
	\setlength{\tabcolsep}{1.35mm}{
		\begin{tabular}{c|c c c}
			\hline \hline
			Methods & S & M & L \\ \hline
			LongShortNet (N=1)                           & 14.2          & 17.3          & 19.7             \\ \hline
			LongShortNet (N=3)                         & 14.6          & 17.5          & 19.8                          \\ \hline
   			DAMO-StreamNet (N=1)                           & 21.0          & 24.2          & 26.2                          \\ \hline
			DAMO-StreamNet (N=3)                           & 21.3          & 24.3          & 26.6                            \\ \hline
	\end{tabular}}
        \vspace{-3mm}
	\label{table:inference-time}
\end{table}

\vspace{-2mm}
\section{Conclusion}
We unveiled DAMO-StreamNet, a meticulously crafted framework, imbibing the latest insights from the YOLO series. DAMO-StreamNet introduces several key innovations, encompassing: (1) the deployment of a solid neck structure fortified with deformable convolution; (2) the origination of a dual-branch structure for mining deeper into time-series data; (3) distillation at the logits stratum; and (4) an advanced real-time forecasting mechanism that contemporaneously refreshes support frame features with the present frame, preparing for the subsequent prediction. Our comparative analyses on the Argoverse-HD dataset evidence DAMO-StreamNet's significant stride over its state-of-the-art counterparts.

\section*{Acknowledgments}
{\small The research work of Zhi-Qi Cheng in this project received support from the US Department of Transportation, Office of the Assistant Secretary for Research and Technology, under the University Transportation Center Program with Federal Grant Number 69A3551747111. Additional support came from the Intel and IBM Fellowships. The views and conclusions contained herein represent those of the authors and not necessarily the official policies or endorsements of the supporting agencies or the U.S. Government.}

{\small
\bibliographystyle{named}
\bibliography{ijcai23}

\begin{thebibliography}{}

\bibitem[\protect\citeauthoryear{Belhassen \bgroup \em et al.\egroup
  }{2019}]{Belhassen_SeqBbox_VISAPP19}
Hatem Belhassen, Heng Zhang, Virginie Fresse, and El{-}Bay Bourennane.
\newblock Improving video object detection by seq-bbox matching.
\newblock In {\em Proceedings of the International Joint Conference on Computer
  Vision, Imaging and Computer Graphics Theory and Applications (VISAPP)},
  pages 226--233, 2019.

\bibitem[\protect\citeauthoryear{Carion \bgroup \em et al.\egroup
  }{2020}]{Carion_DETR_ECCV20}
Nicolas Carion, Francisco Massa, Gabriel Synnaeve, Nicolas Usunier, Alexander
  Kirillov, and Sergey Zagoruyko.
\newblock End-to-end object detection with transformers.
\newblock In Andrea Vedaldi, Horst Bischof, Thomas Brox, and Jan{-}Michael
  Frahm, editors, {\em European Conference on Computer Vision (ECCV)}, pages
  213--229, 2020.

\bibitem[\protect\citeauthoryear{Chen \bgroup \em et al.\egroup
  }{2017}]{Chen_learning_NeurIPS2017}
Guobin Chen, Wongun Choi, Xiang Yu, Tony Han, and Manmohan Chandraker.
\newblock Learning efficient object detection models with knowledge
  distillation.
\newblock {\em Advances in neural information processing systems (NeurIPS)},
  30, 2017.

\bibitem[\protect\citeauthoryear{Chen \bgroup \em et al.\egroup
  }{2018}]{Chen_ScaleTime_CVPR2018}
Kai Chen, Jiaqi Wang, Shuo Yang, Xingcheng Zhang, Yuanjun Xiong, Chen~Change
  Loy, and Dahua Lin.
\newblock Optimizing video object detection via a scale-time lattice.
\newblock In {\em Proceedings of IEEE Conference on Computer Vision and Pattern
  Recognition (CVPR)}, pages 7814--7823, 2018.

\bibitem[\protect\citeauthoryear{Chen \bgroup \em et al.\egroup
  }{2021}]{Chen_SemCKD_AAAI2021}
Defang Chen, Jian-Ping Mei, Yuan Zhang, Can Wang, Zhe Wang, Yan Feng, and Chun
  Chen.
\newblock Cross-layer distillation with semantic calibration.
\newblock In {\em Proceedings of the AAAI Conference on Artificial Intelligence
  (AAAI)}, pages 7028--7036, 2021.

\bibitem[\protect\citeauthoryear{Cheng \bgroup \em et al.\egroup
  }{2018}]{cheng2018learning}
Zhi-Qi Cheng, Xiao Wu, Siyu Huang, Jun-Xiu Li, Alexander~G Hauptmann, and Qiang
  Peng.
\newblock Learning to transfer: Generalizable attribute learning with multitask
  neural model search.
\newblock In {\em Proceedings of the ACM international conference on Multimedia
  (ACM MM)}, pages 90--98, 2018.

\bibitem[\protect\citeauthoryear{Cheng \bgroup \em et al.\egroup
  }{2019a}]{cheng2019learning}
Zhi-Qi Cheng, Jun-Xiu Li, Qi~Dai, Xiao Wu, and Alexander~G Hauptmann.
\newblock Learning spatial awareness to improve crowd counting.
\newblock In {\em Proceedings of the IEEE/CVF international conference on
  computer vision (ICCV)}, pages 6152--6161, 2019.

\bibitem[\protect\citeauthoryear{Cheng \bgroup \em et al.\egroup
  }{2019b}]{cheng2019improving}
Zhi-Qi Cheng, Jun-Xiu Li, Qi~Dai, Xiao Wu, Jun-Yan He, and Alexander~G
  Hauptmann.
\newblock Improving the learning of multi-column convolutional neural network
  for crowd counting.
\newblock In {\em Proceedings of the ACM international conference on multimedia
  (ACM MM)}, pages 1897--1906, 2019.

\bibitem[\protect\citeauthoryear{Cheng \bgroup \em et al.\egroup
  }{2022}]{cheng2022gsrformer}
Zhi-Qi Cheng, Qi~Dai, Siyao Li, Teruko Mitamura, and Alexander Hauptmann.
\newblock Gsrformer: Grounded situation recognition transformer with alternate
  semantic attention refinement.
\newblock In {\em Proceedings of the 30th ACM International Conference on
  Multimedia (ACM MM)}, pages 3272--3281, 2022.

\bibitem[\protect\citeauthoryear{Chin \bgroup \em et al.\egroup
  }{2019}]{Chin_AdaScale_MLSys}
Ting{-}Wu Chin, Ruizhou Ding, and Diana Marculescu.
\newblock Adascale: Towards real-time video object detection using adaptive
  scaling.
\newblock In {\em Proceedings of Machine Learning and Systems (MLSys)}, 2019.

\bibitem[\protect\citeauthoryear{Ding \bgroup \em et al.\egroup
  }{2019}]{Ding_ACNet_ICCV19}
Xiaohan Ding, Yuchen Guo, Guiguang Ding, and Jungong Han.
\newblock Acnet: Strengthening the kernel skeletons for powerful {CNN} via
  asymmetric convolution blocks.
\newblock In {\em International Conference on Computer Vision (ICCV)}, pages
  1911--1920, 2019.

\bibitem[\protect\citeauthoryear{Ding \bgroup \em et al.\egroup
  }{2021a}]{Ding_DBranch_CVPR21}
Xiaohan Ding, Xiangyu Zhang, Jungong Han, and Guiguang Ding.
\newblock Diverse branch block: Building a convolution as an inception-like
  unit.
\newblock In {\em {IEEE} Conference on Computer Vision and Pattern Recognition
  (CVPR)}, pages 10886--10895, 2021.

\bibitem[\protect\citeauthoryear{Ding \bgroup \em et al.\egroup
  }{2021b}]{Ding_RepVGG_CVPR21}
Xiaohan Ding, Xiangyu Zhang, Ningning Ma, Jungong Han, Guiguang Ding, and Jian
  Sun.
\newblock Repvgg: Making vgg-style convnets great again.
\newblock In {\em Proceedings of IEEE Conference on Computer Vision and Pattern
  Recognition (CVPR)}, pages 13733--13742, 2021.

\bibitem[\protect\citeauthoryear{Dosovitskiy \bgroup \em et al.\egroup
  }{2015}]{Dosovitskiy_FlowNet_ICCV15}
Alexey Dosovitskiy, Philipp Fischer, Eddy Ilg, Philip H{\"{a}}usser, Caner
  Hazirbas, Vladimir Golkov, Patrick van~der Smagt, Daniel Cremers, and Thomas
  Brox.
\newblock Flownet: Learning optical flow with convolutional networks.
\newblock In {\em IEEE International Conference on Computer Vision (ICCV)},
  pages 2758--2766, 2015.

\bibitem[\protect\citeauthoryear{Gao \bgroup \em et al.\egroup
  }{2021}]{Gao_VidVRD_MM21}
Kaifeng Gao, Long Chen, Yifeng Huang, and Jun Xiao.
\newblock Video relation detection via tracklet based visual transformer.
\newblock In {\em Proceedings of ACM Conference on Multimedia (ACM MM)}, pages
  4833--4837. {ACM}, 2021.

\bibitem[\protect\citeauthoryear{Ge \bgroup \em et al.\egroup
  }{2021a}]{Ge_OTA_CVPR21}
Zheng Ge, Songtao Liu, Zeming Li, Osamu Yoshie, and Jian Sun.
\newblock {OTA:} optimal transport assignment for object detection.
\newblock In {\em IEEE Conference on Computer Vision and Pattern Recognition
  (CVPR)}, pages 303--312, 2021.

\bibitem[\protect\citeauthoryear{Ge \bgroup \em et al.\egroup
  }{2021b}]{Ge_YOLOX_CORR21}
Zheng Ge, Songtao Liu, Feng Wang, Zeming Li, and Jian Sun.
\newblock {YOLOX:} exceeding {YOLO} series in 2021.
\newblock {\em CoRR}, abs/2107.08430, 2021.

\bibitem[\protect\citeauthoryear{Ghiasi \bgroup \em et al.\egroup
  }{2019}]{GhiasiLL_NASFPN_CVPR19}
Golnaz Ghiasi, Tsung{-}Yi Lin, and Quoc~V. Le.
\newblock {NAS-FPN:} learning scalable feature pyramid architecture for object
  detection.
\newblock In {\em IEEE Conference on Computer Vision and Pattern Recognition
  (CVPR)}, pages 7036--7045, 2019.

\bibitem[\protect\citeauthoryear{Ghosh \bgroup \em et al.\egroup
  }{2021}]{Ghosh_Adaptive_CoRR21}
A.~Ghosh, A.~Nambi, A.~Singh, and et~al.
\newblock Adaptive streaming perception using deep reinforcement learning.
\newblock {\em CoRR}, abs/2106.05665, 2021.

\bibitem[\protect\citeauthoryear{Han \bgroup \em et al.\egroup
  }{2016}]{Han_SeqNMS_CoRR16}
Wei Han, Pooya Khorrami, Tom~Le Paine, Prajit Ramachandran, Mohammad
  Babaeizadeh, Honghui Shi, Jianan Li, Shuicheng Yan, and Thomas~S. Huang.
\newblock Seq-nms for video object detection.
\newblock {\em CoRR}, abs/1602.08465, 2016.

\bibitem[\protect\citeauthoryear{He \bgroup \em et al.\egroup
  }{2021a}]{he2021mgseg}
Jun-Yan He, Shi-Hua Liang, Xiao Wu, Bo~Zhao, and Lei Zhang.
\newblock Mgseg: Multiple granularity-based real-time semantic segmentation
  network.
\newblock {\em IEEE Transactions on Image Processing (TIP)}, 30:7200--7214,
  2021.

\bibitem[\protect\citeauthoryear{He \bgroup \em et al.\egroup
  }{2021b}]{he2021db}
Jun-Yan He, Xiao Wu, Zhi-Qi Cheng, Zhaoquan Yuan, and Yu-Gang Jiang.
\newblock Db-lstm: Densely-connected bi-directional lstm for human action
  recognition.
\newblock {\em Neurocomputing}, 444:319--331, 2021.

\bibitem[\protect\citeauthoryear{Heo \bgroup \em et al.\egroup
  }{2019}]{Heo_Knowledge_AAAI2019}
Byeongho Heo, Minsik Lee, Sangdoo Yun, and Jin~Young Choi.
\newblock Knowledge transfer via distillation of activation boundaries formed
  by hidden neurons.
\newblock In {\em Proceedings of the AAAI Conference on Artificial Intelligence
  (AAAI)}, pages 3779--3787, 2019.

\bibitem[\protect\citeauthoryear{Hinton \bgroup \em et al.\egroup
  }{2015}]{Hinton_distilling_2015}
Geoffrey Hinton, Oriol Vinyals, Jeff Dean, et~al.
\newblock Distilling the knowledge in a neural network.
\newblock {\em arXiv preprint arXiv:1503.02531}, 2(7), 2015.

\bibitem[\protect\citeauthoryear{Huang \bgroup \em et al.\egroup
  }{2018}]{huang2018gnas}
Siyu Huang, Xi~Li, Zhi-Qi Cheng, Zhongfei Zhang, and Alexander Hauptmann.
\newblock Gnas: A greedy neural architecture search method for multi-attribute
  learning.
\newblock In {\em Proceedings of the 26th ACM international conference on
  Multimedia (ACM MM)}, pages 2049--2057, 2018.

\bibitem[\protect\citeauthoryear{Huang \bgroup \em et al.\egroup
  }{2022}]{Huang_TAda_ICLR22}
Ziyuan Huang, Shiwei Zhang, Liang Pan, Zhiwu Qing, Mingqian Tang, Ziwei Liu,
  and Marcelo H.~Ang Jr.
\newblock Tada! temporally-adaptive convolutions for video understanding.
\newblock In {\em Proceedings of International Conference on Learning
  Representations, (ICLR)}, 2022.

\bibitem[\protect\citeauthoryear{Jiang \bgroup \em et al.\egroup
  }{2022}]{Jiang_GiraffeDet_ICLR22}
Yiqi Jiang, Zhiyu Tan, Junyan Wang, Xiuyu Sun, Ming Lin, and Hao Li.
\newblock Giraffedet: {A} heavy-neck paradigm for object detection.
\newblock In {\em International Conference on Learning Representations (ICLR)},
  2022.

\bibitem[\protect\citeauthoryear{Kang \bgroup \em et al.\egroup
  }{2016}]{Kang_TCNN_CVPR16}
Kai Kang, Wanli Ouyang, Hongsheng Li, and Xiaogang Wang.
\newblock Object detection from video tubelets with convolutional neural
  networks.
\newblock In {\em Proceedings of IEEE Conference on Computer Vision and Pattern
  Recognition (CVPR)}, pages 817--825, 2016.

\bibitem[\protect\citeauthoryear{Kim and Lee}{2020}]{Kim_PAA_ECCV20}
Kang Kim and Hee~Seok Lee.
\newblock Probabilistic anchor assignment with iou prediction for object
  detection.
\newblock In Andrea Vedaldi, Horst Bischof, Thomas Brox, and Jan{-}Michael
  Frahm, editors, {\em European Conference on Computer Vision(ECCV)}, volume
  12370, pages 355--371, 2020.

\bibitem[\protect\citeauthoryear{Lan \bgroup \em et al.\egroup
  }{2022}]{lan2022procontext}
Jin-Peng Lan, Zhi-Qi Cheng, Jun-Yan He, Chenyang Li, Bin Luo, Xu~Bao, Wangmeng
  Xiang, Yifeng Geng, and Xuansong Xie.
\newblock Procontext: Exploring progressive context transformer for tracking.
\newblock {\em arXiv preprint arXiv:2210.15511}, 2022.

\bibitem[\protect\citeauthoryear{Li \bgroup \em et al.\egroup
  }{2020}]{LiWR_Streaming_ECCV20}
Mengtian Li, Yu{-}Xiong Wang, and Deva Ramanan.
\newblock Towards streaming perception.
\newblock In {\em Proceedings of the European Conference on Computer Vision
  (ECCV)}, volume 12347, pages 473--488, 2020.

\bibitem[\protect\citeauthoryear{Li \bgroup \em et al.\egroup
  }{2022}]{LongShortNet_CoRR22}
Chenyang Li, Zhi{-}Qi Cheng, Jun{-}Yan He, Pengyu Li, Bin Luo, Han{-}Yuan Chen,
  Yifeng Geng, Jin{-}Peng Lan, and Xuansong Xie.
\newblock Longshortnet: Exploring temporal and semantic features fusion in
  streaming perception.
\newblock {\em CoRR}, abs/2210.15518, 2022.

\bibitem[\protect\citeauthoryear{Lin \bgroup \em et al.\egroup
  }{2014}]{Lin_COCO_ECCV14}
T.~Lin, M.~Maire, S.~Belongie, and et~al.
\newblock Microsoft {COCO:} common objects in context.
\newblock In {\em European Conference on Computer Vision (ECCV)}, volume 8693,
  pages 740--755, 2014.

\bibitem[\protect\citeauthoryear{Lin \bgroup \em et al.\egroup
  }{2017}]{Lin_FPN_CVPR17}
Tsung{-}Yi Lin, Piotr Doll{\'{a}}r, Ross~B. Girshick, Kaiming He, Bharath
  Hariharan, and Serge~J. Belongie.
\newblock Feature pyramid networks for object detection.
\newblock In {\em 2017 {IEEE} Conference on Computer Vision and Pattern
  Recognition (CVPR)}, pages 936--944. {IEEE} Computer Society, 2017.

\bibitem[\protect\citeauthoryear{Lin \bgroup \em et al.\egroup
  }{2020}]{Lin_DualSem_MM20}
Lijian Lin, Haosheng Chen, Honglun Zhang, Jun Liang, Yu~Li, Ying Shan, and
  Hanzi Wang.
\newblock Dual semantic fusion network for video object detection.
\newblock In {\em ACM International Conference on Multimedia (ACM MM)}, pages
  1855--1863, 2020.

\bibitem[\protect\citeauthoryear{Liu \bgroup \em et al.\egroup
  }{2018}]{Liu_PAFPN_CVPR18}
Shu Liu, Lu~Qi, Haifang Qin, Jianping Shi, and Jiaya Jia.
\newblock Path aggregation network for instance segmentation.
\newblock In {\em IEEE Conference on Computer Vision and Pattern Recognition
  (CVPR)}, pages 8759--8768. Computer Vision Foundation / {IEEE} Computer
  Society, 2018.

\bibitem[\protect\citeauthoryear{Liu \bgroup \em et al.\egroup
  }{2020}]{Liu_structured_TPAMI2020}
Yifan Liu, Changyong Shu, Jingdong Wang, and Chunhua Shen.
\newblock Structured knowledge distillation for dense prediction.
\newblock {\em IEEE transactions on pattern analysis and machine intelligence
  (TPAMI)}, 2020.

\bibitem[\protect\citeauthoryear{Mirzadeh \bgroup \em et al.\egroup
  }{2020}]{Mirzadeh_TA_AAAI2020}
Seyed~Iman Mirzadeh, Mehrdad Farajtabar, Ang Li, Nir Levine, Akihiro Matsukawa,
  and Hassan Ghasemzadeh.
\newblock Improved knowledge distillation via teacher assistant.
\newblock In {\em Proceedings of the AAAI conference on artificial intelligence
  (AAAI)}, pages 5191--5198, 2020.

\bibitem[\protect\citeauthoryear{Qiao \bgroup \em et al.\egroup
  }{2022}]{qiao2022real}
Jian-Jun Qiao, Zhi-Qi Cheng, Xiao Wu, Wei Li, and Ji~Zhang.
\newblock Real-time semantic segmentation with parallel multiple views feature
  augmentation.
\newblock In {\em Proceedings of the 30th ACM International Conference on
  Multimedia (ACM MM)}, pages 6300--6308, 2022.

\bibitem[\protect\citeauthoryear{Rezatofighi \bgroup \em et al.\egroup
  }{2019}]{Rezatofighi_GIOU_CVPR19}
Hamid Rezatofighi, Nathan Tsoi, JunYoung Gwak, Amir Sadeghian, Ian~D. Reid, and
  Silvio Savarese.
\newblock Generalized intersection over union: {A} metric and a loss for
  bounding box regression.
\newblock In {\em IEEE Conference on Computer Vision and Pattern Recognition
  (CVPR)}, pages 658--666, 2019.

\bibitem[\protect\citeauthoryear{Sun \bgroup \em et al.\egroup
  }{2021}]{SunHHR21_MAMBA_AAAI}
Guanxiong Sun, Yang Hua, Guosheng Hu, and Neil Robertson.
\newblock {MAMBA:} multi-level aggregation via memory bank for video object
  detection.
\newblock In {\em Proceedings of AAAI Conference on Artificial Intelligence,
  (AAAI)}, pages 2620--2627, 2021.

\bibitem[\protect\citeauthoryear{Tan \bgroup \em et al.\egroup
  }{2020}]{Tan_EfficientDet_CVPR20}
Mingxing Tan, Ruoming Pang, and Quoc~V. Le.
\newblock Efficientdet: Scalable and efficient object detection.
\newblock In {\em 2020 {IEEE/CVF} Conference on Computer Vision and Pattern
  Recognition (CVPR)}, pages 10778--10787, 2020.

\bibitem[\protect\citeauthoryear{Tu \bgroup \em et al.\egroup
  }{2023}]{tu2023implicit}
Shuyuan Tu, Qi~Dai, Zuxuan Wu, Zhi-Qi Cheng, Han Hu, and Yu-Gang Jiang.
\newblock Implicit temporal modeling with learnable alignment for video
  recognition.
\newblock {\em arXiv preprint arXiv:2304.10465}, 2023.

\bibitem[\protect\citeauthoryear{Vasu \bgroup \em et al.\egroup
  }{2022}]{Kumar_Mobile_CoRR22}
Pavan Kumar~Anasosalu Vasu, James Gabriel, Jeff Zhu, Oncel Tuzel, and Anurag
  Ranjan.
\newblock An improved one millisecond mobile backbone.
\newblock {\em CoRR}, abs/2206.04040, 2022.

\bibitem[\protect\citeauthoryear{Wang \bgroup \em et al.\egroup
  }{2018}]{Wang_MANet_ECCV18}
Shiyao Wang, Yucong Zhou, Junjie Yan, and Zhidong Deng.
\newblock Fully motion-aware network for video object detection.
\newblock In Vittorio Ferrari, Martial Hebert, Cristian Sminchisescu, and Yair
  Weiss, editors, {\em Proceedings of European Conference on Computer Vision
  (ECCV)}, volume 11217, pages 557--573, 2018.

\bibitem[\protect\citeauthoryear{Wang \bgroup \em et al.\egroup
  }{2021}]{Wang_HRNet_TPAMI20}
Jingdong Wang, Ke~Sun, Tianheng Cheng, Borui Jiang, Chaorui Deng, Yang Zhao,
  Dong Liu, Yadong Mu, Mingkui Tan, Xinggang Wang, Wenyu Liu, and Bin Xiao.
\newblock Deep high-resolution representation learning for visual recognition.
\newblock {\em IEEE Trans. Pattern Anal. Mach. Intell. (TPAMI)},
  43(10):3349--3364, 2021.

\bibitem[\protect\citeauthoryear{Wang \bgroup \em et al.\egroup
  }{2022}]{Wang_YOLOv7_CORR22}
Chien{-}Yao Wang, Alexey Bochkovskiy, and Hong{-}Yuan~Mark Liao.
\newblock Yolov7: Trainable bag-of-freebies sets new state-of-the-art for
  real-time object detectors.
\newblock {\em CoRR}, abs/2207.02696, 2022.

\bibitem[\protect\citeauthoryear{Xiao and Lee}{2018}]{Xiao_STAligned_ECCV18}
Fanyi Xiao and Yong~Jae Lee.
\newblock Video object detection with an aligned spatial-temporal memory.
\newblock In {\em Proceedings of European Conference on Computer Vision
  (ECCV)}, volume 11212, pages 494--510, 2018.

\bibitem[\protect\citeauthoryear{Yang \bgroup \em et al.\egroup
  }{2022a}]{Yang_Streamyolo_CVPR22}
Jinrong Yang, Songtao Liu, Zeming Li, Xiaoping Li, and Jian Sun.
\newblock Real-time object detection for streaming perception.
\newblock In {\em Proceedings of the Conference on Computer Vision and Pattern
  Recognition (CVPR)}, pages 5385--5395, 2022.

\bibitem[\protect\citeauthoryear{Yang \bgroup \em et al.\egroup
  }{2022b}]{Yang_FGD_CVPR2022}
Zhendong Yang, Zhe Li, Xiaohu Jiang, Yuan Gong, Zehuan Yuan, Danpei Zhao, and
  Chun Yuan.
\newblock Focal and global knowledge distillation for detectors.
\newblock In {\em Proceedings of the IEEE/CVF Conference on Computer Vision and
  Pattern Recognition (CVPR)}, pages 4643--4652, 2022.

\bibitem[\protect\citeauthoryear{Yao \bgroup \em et al.\egroup
  }{2021}]{Yao_Gdetkd_CVPR2021}
Lewei Yao, Renjie Pi, Hang Xu, Wei Zhang, Zhenguo Li, and Tong Zhang.
\newblock G-detkd: Towards general distillation framework for object detectors
  via contrastive and semantic-guided feature imitation.
\newblock In {\em Proceedings of the IEEE/CVF International Conference on
  Computer Vision (ICCV)}, pages 3591--3600, 2021.

\bibitem[\protect\citeauthoryear{Zhu \bgroup \em et al.\egroup
  }{2017a}]{Zhu_Flowguide_ICCV17}
Xizhou Zhu, Yujie Wang, Jifeng Dai, Lu~Yuan, and Yichen Wei.
\newblock Flow-guided feature aggregation for video object detection.
\newblock In {\em Proceedings of IEEE Conference on Computer Vision (ICCV)},
  pages 408--417, 2017.

\bibitem[\protect\citeauthoryear{Zhu \bgroup \em et al.\egroup
  }{2017b}]{Zhu_Deepflow_CVPR17}
Xizhou Zhu, Yuwen Xiong, Jifeng Dai, Lu~Yuan, and Yichen Wei.
\newblock Deep feature flow for video recognition.
\newblock In {\em Proceedings of IEEE Conference on Computer Vision and Pattern
  Recognition (CVPR)}, pages 4141--4150, 2017.

\end{thebibliography}
}
\end{document}